\documentclass[review, times, 10pt, authoryear]{elsarticle}

\usepackage{amssymb}
\usepackage{caption}

\usepackage{lineno}

\usepackage{tabularx}
\usepackage{adjustbox}

\usepackage{algorithmic}
\usepackage{algorithm}
\usepackage{xcolor} 
\usepackage{amsmath}

 \usepackage{threeparttable}

 \usepackage[section]{placeins}

\usepackage{hyperref}

\usepackage{natbib}

\begin{document}

\begin{frontmatter}



\title{Semantic-guided modeling of spatial relation and object co-occurrence for indoor scene recognition}

\author[label1,label2]{Chuanxin Song}
\ead{songchuanxin@mail.sdu.edu.cn}
\author[label1,label2]{Hanbo Wu}
\ead{wuhanbo@sdu.edu.cn}
\author[label1,label2]{Xin Ma\corref{mycorrespondingauthor}}
\ead{maxin@sdu.edu.cn}

\affiliation[label1]{organization={Center for Robotics, School of Control Science and Engineering, Shandong University},
            country={China}}
\affiliation[label2]{organization={Engineering Research Center of Intelligent Unmanned System, Ministry of Education},
            country={China}}
\cortext[mycorrespondingauthor]{Corresponding author}

\begin{abstract}
Exploring the semantic context in scene images is essential for indoor scene recognition. However, due to the diverse intra-class spatial layouts and the coexisting inter-class objects, modeling contextual relationships to adapt various image characteristics is a great challenge. Existing contextual modeling methods for scene recognition exhibit two limitations: 1) They typically model only one type of spatial relationship (order or metric) among objects within scenes, with limited exploration of diverse spatial layouts. 2) They often overlook the differences in coexisting objects across different scenes, suppressing scene recognition performance. To overcome these limitations, we propose SpaCoNet, which simultaneously models Spatial relation and Co-occurrence of objects guided by semantic segmentation. Firstly, the Semantic Spatial Relation Module (SSRM) is constructed to model scene spatial features. With the help of semantic segmentation, this module decouples spatial information from the scene image and thoroughly explores all spatial relationships among objects in an implicit manner, thereby obtaining semantic-based spatial features. Secondly, both spatial features from the SSRM and deep features from the Image Feature Extraction Module are allocated to each object, so as to distinguish the coexisting object across different scenes. Finally, utilizing the discriminative features above, we design a Global-Local Dependency Module to explore the long-range co-occurrence among objects, and further generate a semantic-guided feature representation for indoor scene recognition. Experimental results on three widely used scene datasets demonstrate the effectiveness and generality of the proposed method.

\end{abstract}


\end{frontmatter}


\section{Introduction}
\label{Introduction}
\begin{figure}[htbp]
    \centering
    \includegraphics[width=\textwidth]{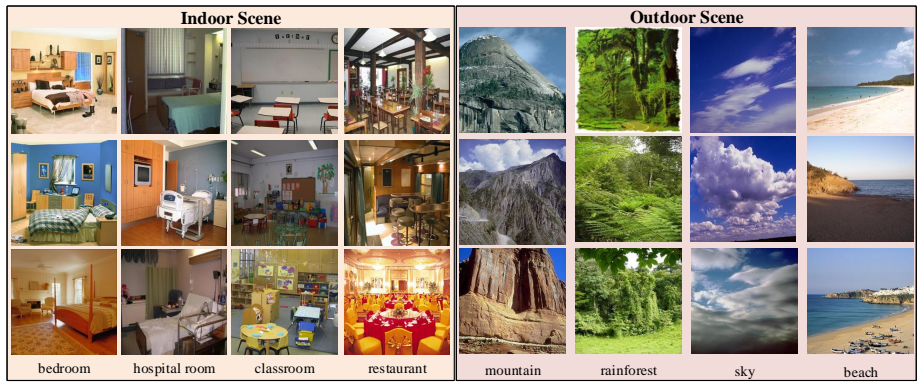}
    \caption{Some examples of different scene datasets. Images in the bedroom and hospital room categories could easily be confused due to shared objects like beds. This inter-class similarity, caused by coexisting objects across scenes, could also be found in the classroom and restaurant (e.g., tables). In contrast, the simple composition makes the variation between the different outdoor scene categories quite obvious.}
    \label{Fig1}
\end{figure}

Scene recognition is a popular research topic in the field of computer vision, aiming to identify the scene categories presented in images, such as "bedroom," "living room," and "kitchen." It is a prerequisite for robots to perform tasks in unknown environments \citep{r36,liu2022service} and plays a key role in various fields such as autonomous driving and human-computer interaction \citep{Outdoor_ESWA,r1}. Indoor scene recognition presents greater challenges compared to its outdoor counterpart, primarily due to the intra-class\footnote{In this paper, the term "class" consistently refers to scene categories.} diverse spatial layout and the coexisting objects in different scene categories, as illustrated in Fig. \ref{Fig1}. Consequently, it is crucial to develop effective methods for indoor scene representation.

Deep Neural Networks (DNN) have been successful in learning advanced representations from images, and they have been widely used in scene recognition tasks in recent years \citep{MVML_ESWA}. However, the performance of DNN for indoor scene recognition is still limited due to the intricate relationships among objects within scenes. Some recent studies\citep{r6} also suggest that scene classification is closely related to the inter-object context it contains. Thus, fully exploiting object information is crucial for recognizing indoor scenes. As shown in Fig. \ref{Fig1}, the varying spatial layouts amplify the intra-class distinctions of indoor scenes, while the presence of coexisting objects in different indoor scenes easily leads to inter-class confusion. Inspired by this phenomenon, we propose SpaCoNet, which aims to simultaneously model \textbf{Spa}tial layout and \textbf{Co}-occurrence of objects for indoor scene recognition.

\begin{figure}[htbp]
    \centering
    \includegraphics[width=0.6\textwidth]{ 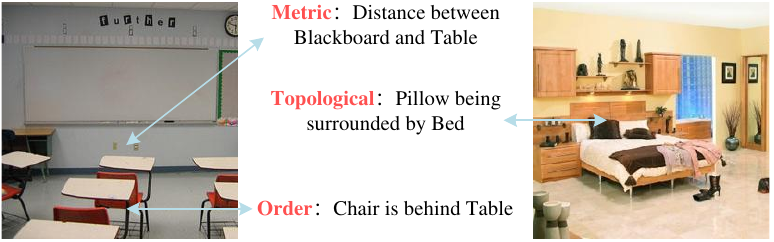}
    \caption{An illustration of three spatial relationships (Metric, Order, Topological) within scenes.}
    \label{Fig_spatial_relation}
\end{figure}

Similar to our insight, several recent strategies have also focused on extracting intra-scene object information for enhancing scene recognition, yielding promising results. To model spatial context, \citep{r37} uses an object-relation-object triplet to explore the spatial order relationship among objects, while \citep{r22,r24} assign object labels to backbone features and model the spatial metric relationship among different object-related features. Despite considerable advancements, these studies focus solely on one type of spatial relationship (i.e., order or metric). In reality, as illustrated in Fig. \ref{Fig_spatial_relation}, spatial relationships encompass not only order and metric types but also topological relationships \citep{spatial1990mathematical}. Relying on any single type of these relationships is insufficient for analyzing such diverse spatial relationships within scenes. Also, these approaches \citep{r22,r24} overlook that space-irrelevant information in backbone features (color, etc.) can also lead to loss degradation during training, which may impede optimization of the network's ability to represent spatial contexts. Considering these limitations, this paper designs a Semantic Spatial Relation Module (SSRM) dedicated to representing spatial contexts. SSRM first decouples the spatial information from scenes through semantic segmentation, ensuring the purity of the network input, thereby thoroughly exploring all spatial relationships among objects implicitly.

In addition to spatial relationships, object co-occurrence across scenes is also a significant contributor to scene recognition. Several approaches \citep{r8,r25,r37} conclude the discriminative objects linked to specific scene categories through statistical analysis of object probability distributions within scenes. However, a significant challenge arises when discriminative objects are identical across different scenes (e.g., the bed in both the bedroom and the hospital room shown in Fig. \ref{Fig1}). This paper proposes a novel idea to solve this challenge inspired by a notable human ability: even if the discriminative objects in the bedroom and the hospital room are identical, humans can still easily discern the differences between these two scenes because the same objects exhibit different characteristics in different scene classes (e.g., apparent variances between the beds in the above two scenes). Building on this observation, before exploring object co-occurrence, we design a Semantic Node Feature Aggregation Module to assign input scene-related features to objects, thus enabling the network to distinguish identical objects like humans. Subsequently, we construct a Global-Local Dependency Module to leverage attention mechanisms to establish long-range co-occurrence among object features, and further generate a semantic-guided feature representation for indoor scene recognition.

In summary, the main contributions of this paper are presented as follows.

\begin{enumerate}
    \item We propose SpaCoNet, a framework that simultaneously models the \textbf{Spa}tial relation and \textbf{Co}-occurrence of objects for indoor scene recognition. Our comprehensive experiments across four different scene datasets (MIT-67 \citep{r11}, SUN397 \citep{r12}, Places \citep{r18}, and reduced SUN RGB-D \citep{song2015sunrgbd}) demonstrate the effectiveness and generality of the proposed method.
    \item A Semantic Spatial Relation Module is designed to model diverse spatial layouts within scenes, which decouples the spatial information from the given image with the help of semantic segmentation, thus avoiding the negative effects of irrelevant information on spatial modeling. By using spatial information as input, this module thoroughly explores all spatial relationships among objects in an implicit manner.
    \item To adequately model object co-occurrence, we first design a Semantic Node Feature Aggregation Module to assign scene-related features to objects, thereby distinguishing identical discriminative objects in different scenes; then construct a Global-Local Dependency Module to explore the long-range co-occurrence among objects using attention mechanisms.
\end{enumerate}

A preliminary version of this work was presented in the conference paper \citep{song2023srrm}. We make significant extensions from various aspects: 1) We introduce a richer type of contextual relationship, i.e., a long-range co-occurrence among objects, to provide more discriminative information for better scene recognition. 2) We optimize the use of spatial relationships among objects within scenes, and evaluate new datasets to demonstrate the generalization of the proposed approach. 3) We conduct comprehensive quantitative ablation studies to show the effectiveness of each new proposed module. 4) A series of qualitative visualization studies for the SpaCoNet are provided to present the characteristics of the proposed method.

\section{Related works}

This section briefly reviews several related researches and examines the differences and connections between related works and our proposed approach.

\subsection{Scene recognition}

\textbf{Handcrafted feature-based methods: }Conventional scene recognition methods usually rely heavily on handcrafted feature extraction. \citep{r9} proposes to use Generalized Search Trees (GIST) to generate a global low-dimensional representation for each image, but ignore the local structure information of the scene. To cope with this problem, some researchers focus on local visual descriptors (such as Local Binary Patterns(LBP) \citep{r2} and Oriented Texture Curves (OTC) \citep{r4}), and use the Bag-of-Visual-Word (BOVW) \citep{r7} to integrate these local visual descriptors into image representation, but do not take spatial structure information into account. For this reason, Spatial Pyramid Matching (SPM) \citep{r10} has been proposed as a component of BOVW, extracting subregions' features and compensating for the missing spatial structure information. Based on the above, Quattoni et al. \citep{r11} propose a prototype-based model for indoor scenes that combines local and global discriminative information. However, the features used by the above methods are handcrafted and low-level, which is limited to distinguish complex scenes.

\textbf{Deep learning-based methods:} In recent years, deep neural networks have made significant progress in computer vision. Several network architectures \citep{r14} have been used to facilitate the development of image classification. Accordingly, many approaches \citep{Efficient_ESWA,MVML_ESWA} attempt to extract visual representations for scene recognition through deep neural networks. For example, Dual CNN-DL \citep{r17} proposes a new dictionary learning layer to replace the traditional FCL and ReLu, which simultaneously enhances features' sparse representation and discriminative ability by determining the optimal dictionary. Lin et al. \citep{r19} propose transforming convolutional features to the ultimate image representation for scene recognition by a hierarchical coding algorithm. These methods utilize convolutional neural networks to extract scene representations, significantly improving the recognition results. However, they still fall far short of the results achieved in tasks such as image classification \citep{r21}. This phenomenon might be due to the lack of effective distinction of coexisting objects within scenes by DNN \citep{r6,r8}. With this in mind, some approaches employ object information in the scene for scene recognition.

\subsection{Semantic segmentation-based modeling}

Semantic segmentation techniques, which can assign semantic labels to each pixel inside an image, have been widely used in diverse applications in recent years. Xu et al. \citep{r41} propose distinguishing the background and foreground through a semantic parsing network, and remedy the raised negative transfer caused by variant background to address the challenge of person reidentification. SGUIE-Net \citep{r42} uses object information as high-level guidance to obtain visually superior underwater image enhancement by region-wise enhancement feature learning. Accordingly, since the classification of a scene is related to the objects it contains, some approaches \citep{song2024semantic, TIP2023Composite} have used the semantic segmentation result to provide auxiliary information for scene recognition. For example, in SAS-Net \citep{r21}, the semantic features generated by the semantic segmentation score tensor are used to add weights to different positions of the feature map generated by the RGB image, so that the network pays more attention to the discriminative regions in the scene image. We will next review object information-based scene recognition methods.

\textbf{Spatial context-based methods:} Considering the large intra-class difference caused by the complex and variable spatial layout within scenes, some studies have explored the spatial relationships between objects for scene recognition. Specifically, Song et al. \citep{r37} propose to explore the spatial order relationship among objects using the "object-relationship-object" triplet, and they encode the triplet as a feature representation based on Recurrent Neural Network (RNN) to assist scene recognition. In addition, ARG-Net \citep{r24} firstly detects the semantic information of regions within scenes through semantic segmentation; then takes the features of different semantic regions as graph nodes and the spatial metric relationship between these regions as graph edges; finally obtains discriminative representations based on the graph convolution for scene recognition. The above studies usually focus on only one spatial relation. Still, as illustrated in Fig. \ref{Fig_spatial_relation}, the spatial relations within scenes are not limited to order and metric relations, but also involve topological relations \citep{spatial1990mathematical}. Existing studies relying solely on one type of spatial relations are insufficient to fully explore the diverse spatial layouts within scenes. In contrast, this paper proposes a Semantic Spatial Relation Module, which takes spatial information directly as input and implicitly models spatial features end-to-end, ensuring thoroughly explores all spatial relations among objects within scenes.

\textbf{Object co-occurrence-based methods:} Besides exploring the spatial context, some approaches \citep{r23} propose using the object co-occurrence across scenes for scene recognition. For example, SDO\citep{r8} proposes to use the co-occurrence probability of objects in different scenes to find representative objects, so that the negative effects of co-occurring objects in multiple scenes can be excluded. Zhou et al. \citep{r25} use the probabilistic method to establish the co-occurrence relationship of objects across scenes, and combine the representative objects with global features to obtain a better scene representation. However, it should be noted that the representative objects in some scenes may be identical, e.g., the bed is representative of both the "bedroom" and the " hospital room" scenes, which poses a significant challenge to the object co-occurrence-based methods described above. In contrast, this paper proposes first to assign scene-related features to objects, and then to model object co-occurrence based on attention mechanisms, avoiding the above challenge.

In summary, semantic segmentation-based scene recognition methods achieve remarkable performance. However, due to the limited precision of semantic segmentation techniques, all semantic segmentation-based methods face the negative impact of semantic ambiguity. This issue is typically mitigated using fixed confidence threshold \citep{r8,r24}, but as the data volume increases, such strategies become inflexible and have limited effectiveness. To address this issue, this paper designs a simple yet effective Adaptive Confidence Filter, adaptively filtering ambiguous points based on each image's specific state, achieving remarkable results.

\subsection{Attention mechanism-based modeling}

The attention mechanism and its variant, the Transformer, have proven effective in modeling long-range dependencies and have been applied across various domains. For instance, SCViT-Net \citep{r40} uses the multi-head self-attention mechanism to model the global interactions of the local structural features for remote sensing image classification. Similarly, Wang et al. \citep{r39} introduce a hybrid CNN-Transformer feature extraction network to combine local details, spatial context, and semantic information for visual place recognition. Additionally, FCT-Net \citep{r38} proposes a Transformer-based framework that combines CNN to improve the discriminative ability of features for scene recognition. 

These methods randomly segment the image into multiple local regions by random cropping or employing a fixed-size grid, and utilize attentional mechanisms to explore inter-region dependencies. However, such random segmentation may result in a single object being dispersed into multiple regions or multiple objects being clustered in a single region, which is not conducive to properly exploring inter-region relationships. In contrast, in this paper, we first obtain the scene-related features corresponding to each object guided by the semantic segmentation label map, thus generating semantic feature sequences. Then, we employ the attention mechanism to explore the long-range co-occurrence among semantic features, resulting in enhanced interpretability.

\section{Our method}
Scene recognition aims to assign predefined scene category labels to given images. Formally, let $I \in {\mathbb{R}^{w \times h \times 3}}$ represent a given image, where $w$ and $h$ denote its width and height, respectively. The set of scene categories to be classified is defined as $T=\left \{ t_1, t_2, ..., t_{\left | T \right |}\right \} $. The objective is to predict the category of the given image $I$ using the following formula:
\begin{equation}
class(I) = \arg \mathop {\max }\limits_{t \in T} (\text{softmax}({\Psi (I)}))_t
\end{equation}
where $\Psi$ represents the designed pipeline for extracting representative information from images (e.g., the SpaCoNet proposed in this paper). In the following, we will first describe the motivation and overview of the proposed SpaCoNet, and then present the specific details of each module.

\subsection{Overview}
\begin{figure}[htbp]
    \centering
    \includegraphics[width=\textwidth]{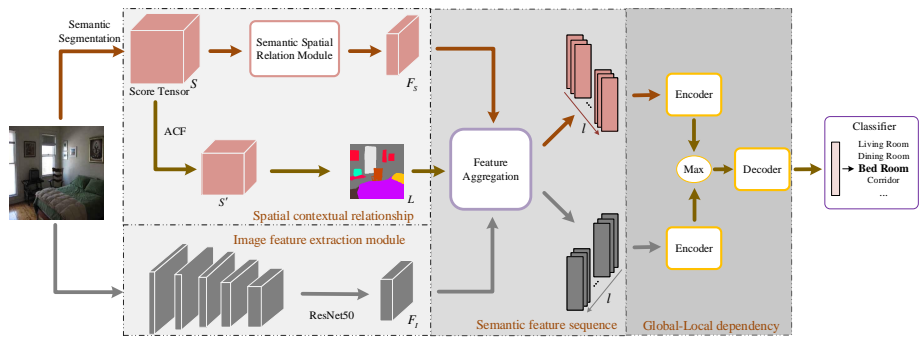}
    \caption{Pipeline of the proposed SpaCoNet for indoor scene representation. Initially, Semantic Spatial Relation Module provides the spatial feature $F_S$, while Image Feature Extraction Module provides the deep feature $F_I$. These features, along with the semantic segmentation label map $L$, are then sent to Semantic Node Feature Aggregation Module, which performs feature aggregation on $F_S$ and $F_I$ guided by $L$, resulting in two semantic feature sequences. Subsequently, Global-Local Dependency Module explores the long-range co-occurrence among semantic features through the attention mechanism, modifying the global features with the obtained information. Finally, the modified features are fed into Classifier to predict the scene category.}
    \label{Fig2}
\end{figure}

This study focuses on indoor scenes captured from a human perspective using digital cameras or smartphones, such as bedrooms, kitchens, and bookstores, etc. As shown in Fig. \ref{Fig1}, indoor scenes are complex, comprising structural elements (e.g., walls, floors, doors), furniture (e.g., sofas, tables, beds), and decorative items (e.g., lamps, pictures, plants). Overall, the intra-scene diverse spatial layout and the coexisting objects across scenes present significant challenges for indoor scene recognition. To meet these challenges, as shown in Fig. \ref{Fig2}, this paper proposes SpaCoNet, guided by semantic information, aiming to simultaneously model the \textbf{Spa}tial layout and \textbf{Co}-occurrence of objects to extract representative information for indoor scene recognition.

Initially, we employ PlacesCNN \citep{r18} as the Image Feature Extraction Module (IFEM) to extract deep scene features, which can be regarded as the baseline. Next, considering the diverse spatial layouts, we construct a Semantic Spatial Relation Module (SSRM) dedicated to modeling spatial features within the scene. Previous studies \citep{r22,r24} tend to explore one type of spatial relation, which cannot fully explore the diverse spatial layouts within scenes. In contrast, SSRM takes spatial information directly as input and implicitly models spatial features end-to-end, ensuring thoroughly explores all spatial relations among objects.

Furthermore, considering the inter-class similarity caused by coexisting objects across scenes, we further model the object co-occurrence to assist scene recognition. Previous studies \citep{r8,r25} typically identify representative objects for scene categories from a statistical perspective. However, representative objects may be identical in some scenes, which poses a significant challenge to the above studies. To overcome this challenge, we design a Semantic Node Feature Aggregation Module to assign the input scene-related features (features extracted by IFEM and SSRM) to each object node. In this way, scene-related features are assigned to objects, thereby distinguishing objects with the same semantics in different scenes, and generating semantic feature sequences.

Based on the above, given that objects with the same semantics exhibit different features in different scenes, the traditional statistical strategy \citep{r8} is no longer suitable for exploring object co-occurrence in this work. Fortunately, the advent of Transformer \citep{r33} provides a flexible architecture with a multi-head attention mechanism, which can capture the long-range dependencies among sequential features. Therefore, we construct a Global-Local Dependency Module to deeply investigate the long-range correlation among semantic features and global features through attention mechanisms, fine-tuning the global features while modeling object co-occurrence to obtain a more discriminative scene representation.

Overall, to simultaneously model the Spatial layout and Co-occurrence of objects, our proposed SpaCoNet contains five modules. Initially, to thoroughly explore diverse spatial relations within scenes, we design the \textbf{(a) Semantic Spatial Relation Module}, which implicitly models spatial features end-to-end. Concurrently, \textbf{(b) Image Feature Extraction Module} processes the original image to fully utilize scene information. Furthermore, to address the challenge of modeling object co-occurrence in scenes where representative objects may be identical, the feature maps from (a) and (b) are passed to \textbf{(c) Semantic Node Feature Aggregation Module}. (c) assign input scene-related features to each object node, thereby distinguishing objects with identical semantics in different scenes, and generating semantic feature sequences. These sequences are then forwarded to\textbf{ (d) Global-Local Dependency Module}, which explores long-range co-occurrence within semantic feature sequences through the attention mechanism and modifies the global features with the obtained information. Finally, the modified features are fed into \textbf{(e) Scene Recognition Module} to predict the scene category.

\subsection{Semantic Spatial Relation Module}

Spatial relationships, as discussed in \citep{spatial1990mathematical}, encompass three types: topological relationships (e.g., a pillow being surrounded by a bed), order relationships (e.g., a chair is behind a table), and metric relationships (e.g., the distance between an artboard and a person). Relying solely on one type of relationship (Order \citep{r37} or Metric \citep{r24}) is insufficient for analyzing such diverse spatial layouts. Therefore, we construct the Semantic Spatial Relation module (SSRM) as a dedicated branch to model spatial relationships in an implicit manner.

\begin{figure}[htbp]
    \centering
    \includegraphics[width=0.75\textwidth]{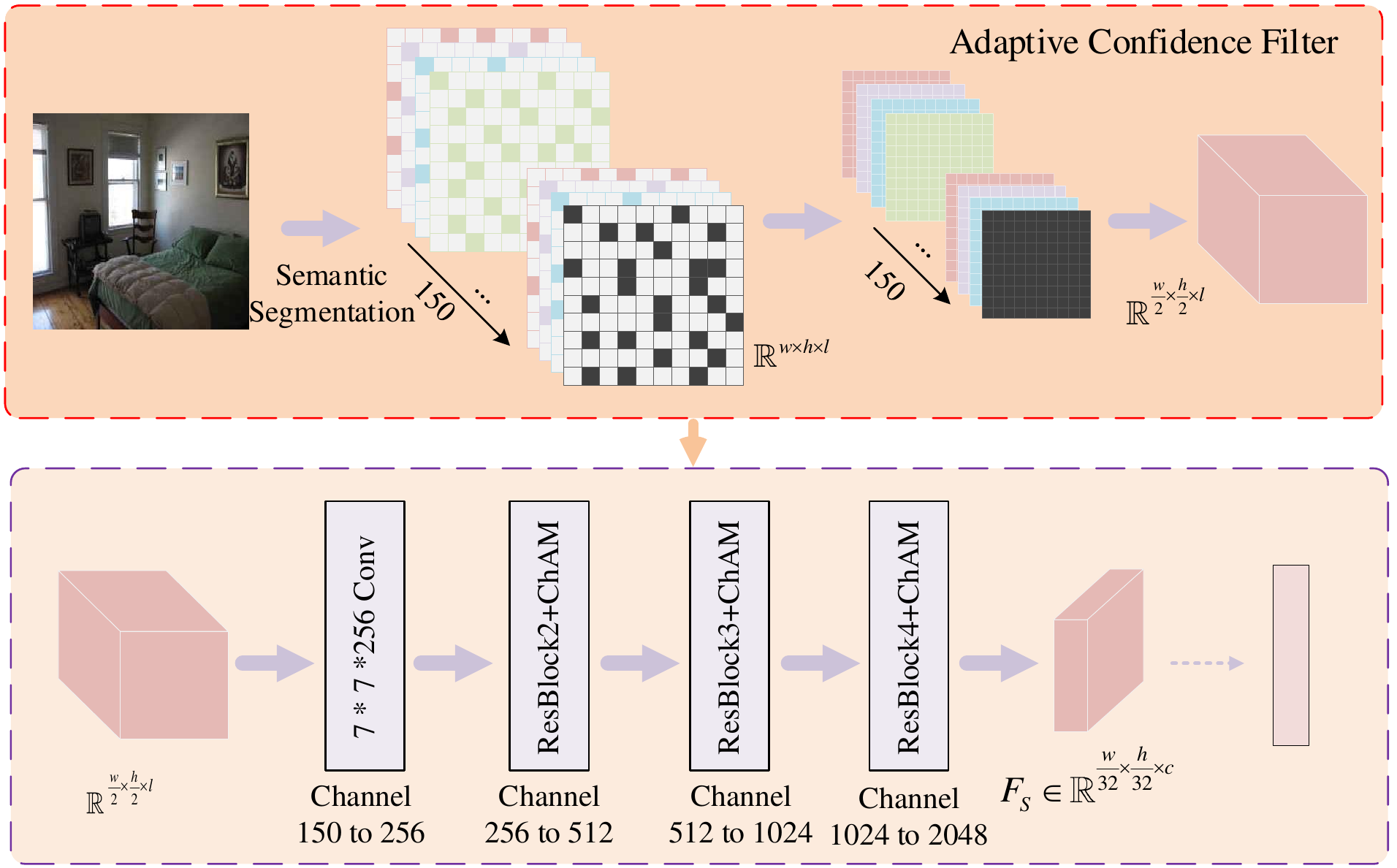}
    \caption{Semantic Spatial Relation Module (SSRM). The part surrounded by the red dashed box represents the confidence filtering stage, which is used to address semantic ambiguities. The part surrounded by the purple dashed box represents the spatial context modeling stage, which is used to implicitly model spatial features from the provided spatial information.}
    \label{Fig3}
\end{figure}

As stated in Section \ref{Introduction}, space-irrelevant information (such as color) within scenes can expedite loss reduction during training. However, this reduction is not conducive to optimizing the network's ability to model spatial context, potentially leading to sub-optimal performance in processing spatial information. Since the objective of SSRM is to thoroughly explore spatial information within scenes, we decouple the pure spatial information from input scenes. This pure spatial information then serves as the input for this module, avoiding the negative impact of space-irrelevant information during training. Fig. \ref{Fig3} illustrates the framework of the proposed SSRM.

To be specific, the given image $I \in {\mathbb{R}^{w \times h \times 3}}$ is first processed through a semantic segmentation network to obtain a score tensor $S \in {\mathbb{R}^{w \times h \times l}}$($l=150$) \citep{r27,r28}. This tensor focuses solely on pure spatial information, filtering out non-essential elements. Then, we use $S$ as the input for SSRM, ensuring that the module is devoid of space-irrelevant information and can fully explore the spatial context. The SSRM generates a spatial context feature ${F_S} \in {\mathbb{R}^{\frac{w}{{32}} \times \frac{h}{{32}} \times c}}$, which represents the concentrated representation of the spatial contextual information in $I$. Based on the ResNet50 \citep{r14}, we design a more suitable architecture for SSRM according to the specific characteristics of the segmentation score tensor. Compared with the original ResNet50, SSRM is more efficient and requires less computing power. 

Due to the limited precision of the semantic segmentation network, errors in segmentation results are inevitable. Several existing methods \citep{r8,r24} have attempted to address this issue by implementing fixed confidence thresholds to filter ambiguous points. However, these methods are inflexible and have limited efficacy, especially when dealing with large datasets. In contrast, to alleviate the adverse effects caused by semantic ambiguity, we design an Adaptive Confidence Filter (ACF) based on max pooling. The ACF is capable of dynamically adapting to the state of the image, allowing for flexible filtering of each semantic segmentation score tensor. 

\begin{figure}[htbp]
    \centering
    \includegraphics[width=0.9\textwidth]{ 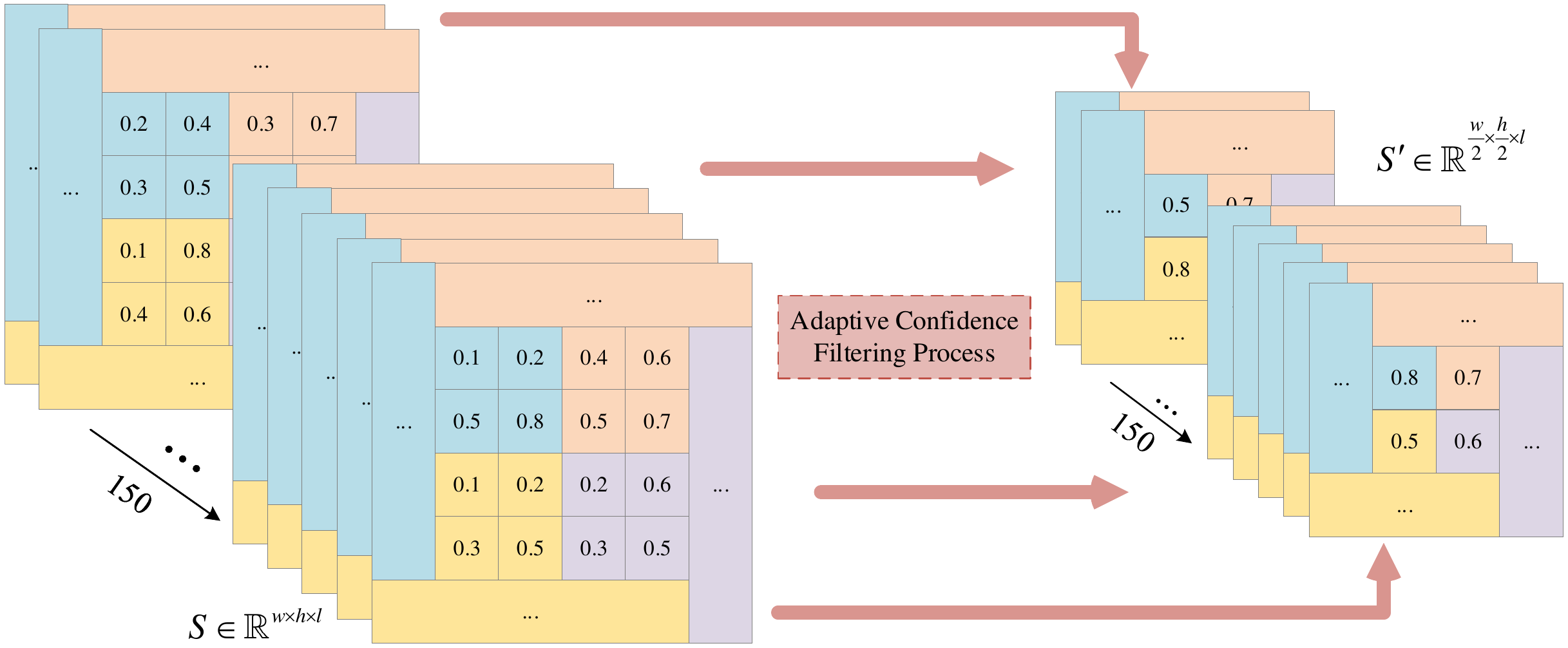}
    \caption{An example of the proposed Adaptive Confidence Filter (ACF).}
    \label{Fig4}
\end{figure}

Specifically, ACF employs a $2 \times 2$ max pooling filter to smooth each channel of $S$. For each domain the filter is applied to, ACF selects only the pixels with the highest confidence in its coverage, and generates the output $S' \in {\mathbb{R}^{\frac{w}{2} \times \frac{h}{2} \times l}}$ after processing all the channels of $S$. Fig. \ref{Fig4} illustrates an example of ACF. Compared to the segmentation map corresponding to $S$, in which internal points represent channels with the highest confidence within the $1 \times 1 \times l$ range, each pixel point in the segmentation map corresponding to $S'$ represents the channel with the highest confidence within the $2 \times 2 \times l$ range in which it is located. By leveraging the coverage of the discriminative domain instead of a fixed threshold, ACF filters the semantic segmentation map, enabling it to adjust to the unique characteristics of each image. Consequently, ACF improves the precision and generalizability of the semantic segmentation map. Moreover, ACF reduces the subsequent networks' input size, which reduces the computational cost of SSRM. We provide a comparative analysis of this concept in Section \ref{SSRM_eval}.

Next, the filtered semantic segmentation score tensor is processed using ResBlocks to explore spatial relationships among object regions. ResBlocks includes $Basic Blocks$ 2, 3, and 4 of the original ResNet50. Inspired by the fact that each channel value in ${S'_{(i,j)}}$ represents the predicted semantic probability of the corresponding pixel in image $I$ \citep{r28}, following \citep{r21}, the Channel Attention Module (ChAM) within CBAM \citep{r26} is introduced between ResBlocks to help the network to focus better on the critical semantic categories in the image. 

Through exploring the spatial relation among object regions within $S'$, this module can generate the semantic-based spatial feature $F_S$, which implicitly encompasses comprehensive spatial context information.

\subsection{Image Feature Extraction Module}

In addition to spatial relationships, object co-occurrence across scenes is also a significant contributor to scene recognition. Of course, spatial relationships also contain shallow multi-object co-occurrence information. Still, their focus is on exploring the positional relationships among objects,  irrespective of their intrinsic properties, which is also why we use the decoupled $S$ as the input to the SSRM to explore spatial relations. However, since $S$ does not contain information about the objects' characteristics (color, texture, etc.), the SSRM lacks the ability to distinguish discriminative objects that may appear in different scenes, leading to under-exploration of object co-occurrence. Therefore, we introduce an Image Feature Extraction Module (IFEM) for $I$ to extract the complete image deep feature $F_I$.

As shown in Fig. \ref{Fig2}, the IFEM comprises a nearly complete ResNet50 architecture. Owing to its pretraining on the Places dataset \citep{r18}, following \citep{r25,song2023srrm}, we also refer to the IFEM as PlacesCNN. Notably, we exclude the pooling layer from the ResNet50 since it causes translation invariance, leading to the blurring of distinctions between various node features in the top-level semantic feature layer \citep{r35}. This blurring hinders the exploration of the global-local dependencies that follow. Therefore, we remove it to enhance the model's performance in identifying and distinguishing such dependencies. Given an image $I \in {\mathbb{R}^{w \times h \times 3}}$, IFEM generates an advanced deep feature ${F_I} \in {\mathbb{R}^{\frac{w}{{16}} \times \frac{h}{{16}} \times c}}$. Also, since the input of IFEM is image $I$, we perform global average pooling on its output $F_I$, and then the scene classification results obtained through a fully connected layer can be used as a baseline for this study.

 After the above process, utilizing $F_I$ and $F_S$, we can link scene-related features with objects in the subsequent Semantic Node Feature Aggregation Module, so as to differentiate the same objects in different scenes, laying a solid foundation for fully exploring the correlation between objects within scenes. Note that for tractability, we interpolate $F_S$ to the same resolutions as $F_I$ using the Bilinear interpolation algorithm.

\subsection{Semantic Node Feature Aggregation Module}

To differentiate the same objects across various scenes, we design the Semantic Node Feature Aggregation Module, which allocates objects with input scene-related features. By successively feeding spatial features $F_S$ and deep features $F_I$ to this module, we can obtain their respective semantic feature sequences.

For each image, we first obtain its semantic segmentation score tensor $S$ from the SSRM, and generate the label map $L$ from the probabilistic relationships within $S$. $L$ enables us to extract the object information corresponding to each point in $F_S$ and $F_I$. However, two issues arose:

\begin{enumerate}
    \item Semantic ambiguity issue: Due to the limited precision of semantic segmentation, there can be errors in $L$, which may negatively affect subsequent works.
    \item Resolution mismatch issue: The resolution of $L$ is the same as the resolution of the input image, which is different from the resolution of the feature map.
\end{enumerate}

\begin{figure}[htbp]
    \centering
    \includegraphics[width=\textwidth]{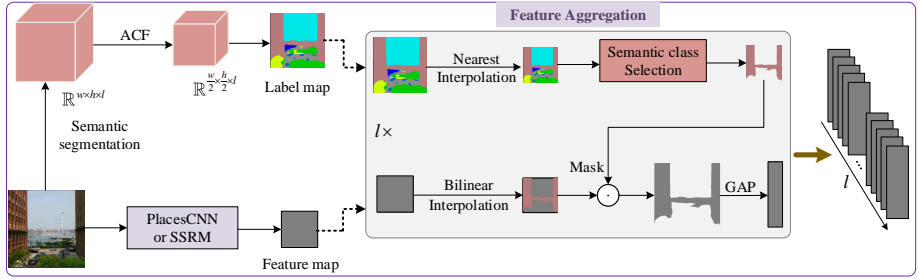}
    \caption{The aggregation process of the Semantic Node Feature Aggregation Module for image features or spatial features. Note that this module handles these two feature maps separately.}
    \label{Fig5}
\end{figure}

To address these issues, we perform feature aggregation using the strategy described below. Fig. \ref{Fig5} illustrates the framework of this module, while the corresponding pseudocode is detailed in Algorithm \ref{alg:alg1}.

\begin{algorithm}[htb]
\caption{Algorithm for generating semantic feature sequences}\label{alg:alg1}
\textbf{Input}: Spatial feature $F_S$, Deep feature $F_I$, Semantic segmentation score tensor $S$;\\
\textbf{Output}: Feature sequence $X_{spa} \in {\mathbb{R}^{l \times c}}$ and $X_{rgb} \in {\mathbb{R}^{l \times c}}$;
\begin{algorithmic}[1] 
\STATE $S' = ACF (S)$; \# Filtering semantic ambiguity using Adaptive Confidence Filter
\STATE $L' = Argmax (S')$; \# Generate semantic segmentation label map $L'$ for $S'$
\STATE $L'$ = $Nearest Neighbor Interpolation (L')$; \# $L' \in {\mathbb{R}^{\frac{w}{16} \times \frac{h}{16}}}$
\FOR{$o=0$ to $l-1$}
    \FOR{$i=1$ to $\frac{w}{16}$} 
        \FOR{$j=1$ to $\frac{h}{16}$}
            \STATE $L_{i, j}^{o}=\left\{\begin{array}{ll}1, & L_{i, j}^{\prime}=o \\0, & L_{i, j}^{\prime} \neq o\end{array}\right.$; \# Generate a binary map $L_o$ for object $o$
        \ENDFOR
    \ENDFOR
    \STATE $\begin{array}{l}    x_{{rgb}}^{o}={ AveragePooling }\left(F_{I} \odot L^{o}\right) \\    x_{{spa }}^{o}={ AveragePooling }\left(F_{S} \odot L^{o}\right)\end{array}$; \# Generate scene-related features for $o$
\ENDFOR
\STATE $\begin{array}{l}  {X_{rgb}} = concat(x_{rgb}^0,x_{rgb}^1,x_{rgb}^2,...,x_{rgb}^{l-1}) \hfill \\  {X_{spa}} = concat(x_{spa}^0,x_{spa}^1,x_{spa}^2,...,x_{spa}^{l-1}) \hfill \\ \end{array}$;
\STATE \textbf{return} $X_{spa} \in {\mathbb{R}^{l \times c}}$, $X_{rgb} \in {\mathbb{R}^{l \times c}}$;
\end{algorithmic}
\end{algorithm}

For the Semantic ambiguity issue, we first apply our ACF on the score tensor $S$ to obtain $S'$. Subsequently, the refined tensor is used to generate the label map $L' \in {\mathbb{R}^{\frac{w}{2} \times \frac{h}{2}}}$. In this way, we enhance the confidence level of the segmentation results. 

For the Resolution mismatch issue, after obtaining $L'$, we use the nearest-neighbor interpolation algorithm to reduce its resolution to be consistent with the feature map $F_I$ and $F_S$.

Subsequently, for each object $o$ in $l$ semantic categories, we generate a binary map ${L^o}$, as follows:
\begin{equation}
 L_{i, j}^{o}=\left\{\begin{array}{ll}
1, & L_{i, j}^{\prime}=o \\
0, & L_{i, j}^{\prime} \neq o
\end{array}\right.
\end{equation}

We apply the binary map ${L^o}$ over the feature map $F_I$ and $F_S$, respectively, to extract the image features and spatial features related to object $o$. Then, we perform average pooling to obtain the deep feature vector $x_{rgb}^o \in {\mathbb{R}^c}$ and the spatial feature vector $x_{spa}^o \in {\mathbb{R}^c}$ of object $o$, which can be expressed as:
\begin{equation}
\begin{array}{l}
    x_{{rgb}}^{o}={ AveragePooling }\left(F_{I} \odot L^{o}\right) \\
    x_{{spa }}^{o}={ AveragePooling }\left(F_{S} \odot L^{o}\right)
\end{array}
\end{equation}
where $ \odot $ represents the Hadamard product.

Finally, the feature vectors stack together to form the deep aggregation feature ${X_{rgb}} \in {\mathbb{R}^{l \times c}}$ and the spatial aggregation feature ${X_{spa}} \in {\mathbb{R}^{l \times c}}$. The above process can be formulated as follows:
\begin{equation}
\begin{array}{l}
  {X_{rgb}} = concat(x_{rgb}^0,x_{rgb}^1,x_{rgb}^2,...,x_{rgb}^{149}) \hfill \\
  {X_{spa}} = concat(x_{spa}^0,x_{spa}^1,x_{spa}^2,...,x_{spa}^{149}) \hfill \\ 
\end{array}
\end{equation}

Both $X_{rgb}$ and $X_{spa}$ are then fed into the Global-Local Dependency Module to explore the object co-occurrence within semantic feature sequences.

\subsection{Global-Local Dependency Module}

\begin{figure}[htbp]
    \centering
    \includegraphics[width=\textwidth]{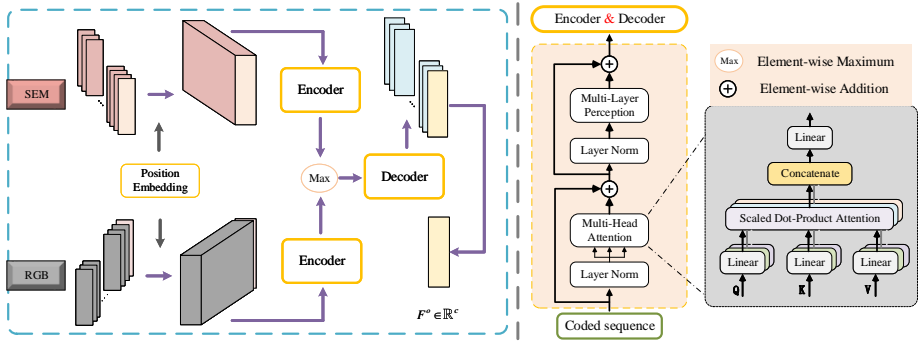}
    \caption{Detailed structure of Global-Local Dependency Module. First, two Encoders explore the internal correlations within $X_{rgb}$ and $X_{spa}$, respectively. Then, these two features are merged, followed by a Decoder to explore the intrinsic relationship of the overall information.}
    \label{Fig6}
\end{figure}

Given that the same object has different characteristics in various scenes, manual strategies are no longer appropriate for exploring the object co-occurrence in the proposed framework. To address this problem, we construct the Global-Local Dependency Module, establishing the long-range correlation among semantic node features utilizing the attention mechanism. This correlation can then be used to modify the global feature representation. The overall process is shown in Fig. \ref{Fig6}. Considering the disparities between feature sequences $X_{rgb}$ and $X_{spa}$, we design an Encoder-Decoder architecture, i.e., we first explore their internal correlation individually, then merge them to explore the intrinsic relationship of the overall information.

To be specific, from the semantic node feature aggregation module, we obtain the deep feature sequence ${X_{rgb}} \in {\mathbb{R}^{l \times c}}$ and the spatial feature sequence ${X_{spa}} \in {\mathbb{R}^{l \times c}}$. Since our objective is to use the long-range correlation among semantic features to modify the global representation, we make changes to these two feature sequences by using their respective global features as nodes, which are obtained through global average pooling on $F_I$ and $F_S$, resulting in extended feature sequences $X_{rgb}^1 \in {\mathbb{R}^{(l + 1) \times c}}$ and $X_{spa}^1 \in {\mathbb{R}^{(l + 1) \times c}}$:

\begin{equation}
\begin{array}{l}
  X_{rgb}^1 = concat({X_{rgb}},GlobalAvgPooling({F_I})) \hfill \\
  X_{spa}^1 = concat({X_{spa}},GlobalAvgPooling({F_S})) \hfill \\ 
\end{array}
\end{equation}

Meanwhile, considering the necessity of recording object categories corresponding to each semantic feature, position embedding $P_{emb}^{rgb} \in {\mathbb{R}^{(l + 1) \times c}}$ and $P_{emb}^{spa} \in {\mathbb{R}^{(l + 1) \times c}}$ are further added to generate coded feature sequences $X_{spa}^2 \in {\mathbb{R}^{(l + 1) \times c}}$ and $X_{rgb}^2 \in {\mathbb{R}^{(l + 1) \times c}}$. Hence, the inputs of two Encoders are formulated as follows:
\begin{equation}
\begin{array}{l}
  X_{rgb}^2 = X_{rgb}^1 + P_{emb}^{rgb} \hfill \\
  X_{spa}^2 = X_{spa}^1 + P_{emb}^{spa} \hfill \\ 
\end{array} 
\end{equation}

Then, we employ the attention mechanism to explore the long-range dependencies among local semantic node features and between them and the global node feature. Inspired by Vision Transformer \citep{r33}, each Encoder block consists of three main components: Layer normalization, Multi-head Self-Attention (MSA), and Multi-Layer Perceptron (MLP). Layer normalization is used to normalize and smooth data distribution, thereby improving the model's generalization ability. The MSA follows it. Next, a residual connection is employed to convey information to alleviate overfitting. Afterward, Layer normalization is applied again, and the output of the Encoder block is obtained through MLP. The entire operation of the Encoder for processing the image aggregation feature $X_{rgb}^2 \in {\mathbb{R}^{(l + 1) \times c}}$ and the spatial aggregation feature $X_{spa}^2 \in {\mathbb{R}^{(l + 1) \times c}}$ is represented by follows:
\begin{equation}
\begin{array}{l}
  X_{rgb}^3 = X_{rgb}^2 + MSA(LN(X_{rgb}^2)) \hfill \\
  X_{rgb}^4 = X_{rgb}^3 + MLP(LN(X_{rgb}^3)) \hfill \\ 
  X_{spa}^3 = X_{spa}^2 + MSA(LN(X_{spa}^2)) \hfill \\
  X_{spa}^4 = X_{spa}^3 + MLP(LN(X_{spa}^3)) \hfill \\ 
\end{array} 
\end{equation}

With two Encoders, we individually explore the long-range dependencies among the internal node features of $X_{rgb}^2$ and $X_{spa}^2$. Subsequently, we merge them by taking the element-wise Maximum:
\begin{equation}
\label{eq_maximum}
{X_o} = Max(X_{spa}^4,X_{rgb}^4)
\end{equation}

To better utilize the long-range dependency among node features for scene representation, the output $X_o$ is fed into the Decoder to mitigate the disparities between image deep features and spatial features, and to explore the overall information's intrinsic relationships. The structure of the Decoder is similar to that of the Encoder and can be formulated as follows:
\begin{equation}
\begin{array}{l}
  X_o^1 = {X_o} + MSA(LN({X_o})) \hfill \\
  X_o^2 = X_o^1 + MLP(LN(X_o^1)) \hfill \\ 
\end{array} 
\end{equation}

With the implementation of the Encoder and Decoder, we establish a robust long-range correlation among all node features. This integration process generates an optimized representation $X_o^2$.

\subsection{Scene Recognition Module}

To optimize the scene representation, we employ attention mechanisms (Encoder and Decoder) to model the long-range dependency between global node and local nodes. To prevent over-fitting, we only adopt a one-layer Encoder and one-layer Decoder. After the global-local dependency modeling, we obtain the optimized representation $X_o^2$. In classification processing, we extract the fully modified global node feature $F_o \in {\mathbb{R}^{c}}$ from $X_o^2$ as the final global representation (as shown in Fig. \ref{Fig6}). These representations are fed into a fully connected network to obtain the final scene prediction, and the cross entropy function is used to compute the final loss:
\begin{equation}
    L =  - \sum\limits_{n = 1}^N {{y_n}\log \frac{{{exp({F_o}(n))}}}{{\sum\nolimits_{m = 1}^N {{exp({F_o}(m))}} }}} 
\end{equation}
where $y$ is the ground truth and $F_o$ is the output of this module.

\section{Experiment}

This section begins by presenting the benchmark datasets and performing ablation experiments to assess the impact of each module on the proposed method. Subsequently, we compare the proposed method with the state-of-the-art methods.

\subsection{Datasets}

\textbf{MIT-67 Dataset} \citep{r11} consists of 67 indoor scene classes with a total of 15620 images, and each scene category contains at least 100 images. Following the recommendations by the authors \citep{r11}, each class has 80 images for training and 20 images for testing. The evaluation of the MIT dataset is challenging due to the large intra-class variation of indoor scenes.

\textbf{SUN397 Dataset} \citep{r12} is a large dataset covering indoor and outdoor scenes. It contains 397 scene categories spanning 175 indoor and 220 outdoor scene categories, each containing at least 100 RGB images. This study focuses on the 175 indoor scene categories to evaluate our proposed approach. Following the evaluation protocol of the original paper \citep{r12}, we randomly select 50 images from each scene class for training and another 50 for testing.

\textbf{Places Dataset} \citep{r18} is one of the largest scene recognition datasets, which includes about 1.8 million training images and 365 scene categories. This paper uses a simplified version, and only indoor scene categories are considered. To ensure a fair comparison with other indoor scene recognition methods\citep{r25}, we used the same two scene class settings as them, namely Places\_7 and Places\_14. Places\_7 contains seven indoor scenes: Bath, Bedroom, Corridor, Dining Room, Kitchen, Living Room, and Office. Places\_14 contains 14 indoor scenes: Balcony, Bedroom, Dining Room, Home Office, Kitchen, Living Room, Staircase, Bathroom, Closet, Garage, Home Theater, Laundromat, Playroom, and Wet Bar. For the test set, we use the same setup as the official dataset \citep{r18}.

\textbf{SUN RGB-D Dataset} \citep{song2015sunrgbd} is a comprehensive RGB-D dataset. It comprises 3874 Microsoft Kinect v2 images, 3389 Asus Xtion images, 2003 Microsoft Kinect v1 images and 1159 Intel RealSense images. The diversity of categories and sources makes SUN RGB-D more suitable for verifying the algorithm's generalization. Following \citep{r25,song2023srrm}, we used the reduced SUN RGB-D Dataset, focusing on the same classes as Places\_7, to test our model pretrained on Places\_7 without any fine-tuning on the SUN RGB-D dataset.

\subsection{Implementation details}
Vision Transformer Adapter\citep{r27} that is pretrained on ADE20K dataset \citep{r28} is used as the semantic segmentation network. Given an image $I \in {\mathbb{R}^{w \times h \times 3}}$, its output is a score tensor $S \in {\mathbb{R}^{w \times h \times l}}$. This tensor can be used to generate the semantic label map  $L \in {\mathbb{R}^{w \times h}}$ for $I$. The semantic prediction probability of location $(i,j)$ in $I$ is denoted by ${S_{i,j}} \in {\mathbb{R}^{1 \times 1 \times l}}$, where $l$ represents the number of semantic labels ($l=150$). In $L$,  each pixel $(i,j)$ is assigned a value ${L_{ij}}$, representing the semantic label of its corresponding pixel in the input image.

To thoroughly explore the spatial information within the input scene, it is essential to eliminate the impact of spatially irrelevant information on the network parameters when training the SSRM. As a result, a two-stage training procedure is implemented. In the first stage, both the SSRM and IFEM are augmented with a global average pooling layer and a classifier to fine-tune the two pretrained modules separately. In the second stage, we freeze the weights of SSRM and IFEM and train only the subsequent modules. To ease the training process, we use ALI-G \citep{r32} to optimize the network parameters in both stages. ALI-G requires only an initial learning rate hyperparameter, which is set to 0.01 in the first stage and 0.1 in the second stage. To mitigate overfitting, we implement dropout regularization in our classifier, applying dropout rates of 0.3 and 0.8 in the first and second stages, respectively. During both training stages, the batch size is set to 32 for the MIT-67 and SUN397 datasets, while for the larger and quicker-converging Places dataset, the batch size is increased to 64.

Consistent with previous studies \citep{DeepScene_ESWA,EMLELM_ESWA}, recognition accuracy is measured using top-1 accuracy, which represents the percentage of validation/testing images where the top-scored class matches the ground-truth label:
 \begin{equation}
Accuracy = \frac{{{N_C}}}{{{N_T}}}
\end{equation}
where $N_C$ is the number of correctly categorized images, and $N_T$ is the total number of images.

 All experiments are conducted on a single NVIDIA 3090 GPU using the PyTorch \citep{paszke2019pytorch} and SAS-Net \citep{r21} open-source framework. When evaluating the final performance, the standard 10-crop testing method \citep{r16} is used for comparison with other methods.

\subsection{Quantitative ablation analysis}
In this subsection, we conduct a series of quantitative ablation studies to evaluate the effectiveness of the proposed method.

\subsubsection{Evaluation of the Semantic Spatial Relation Module}
\label{SSRM_eval}

In this part, we evaluate the proposed SSRM on the MIT-67 and SUN397 datasets, and study the effect of the Adaptive Confidence Filter (ACF) and Channel Attention module (ChAM) on the recognition performance. The experimental results are presented in Table \ref{tab1}.

\begin{table}[htbp]
    \footnotesize
    \centering
    \caption{Ablation results for different architectures for the SSRM (\%).}
    \begin{tabular}{lcccc}
    \hline
        Architecture & Pretraining & MIT-67 & SUN397 & Flops (G) \\ \hline
        ResNet50 & Scratch & 64.403 & 50.529 & 27.2  \\ 
        $4 \times 4$ ACF + ResNet50 & Scratch & 70.149 & 52.894 & 9.31  \\ 
        $2 \times 2$ ACF + ResNet50 & Scratch & 69.627 & 53.035 & 9.31  \\ 
        ResNet50\_ChAM & Scratch & 69.851 & 54.624 & 27.3  \\ 
        $4 \times 4$ ACF + ResNet50\_ChAM & Scratch & 73.284 & 57.4 & 9.32  \\ 
        $2 \times 2$ ACF + ResNet50\_ChAM & Scratch & \textbf{74.403} & \textbf{58.329} & 9.32  \\ \hline
        \# $2 \times 2$ ACF +ResNet50\_ChAM & Places & 81.642 & 66.953 & 9.32  \\ \hline
        \multicolumn{4}{l}{\# indicates that the model's parameters are pretrained on the Places dataset.}
    \end{tabular}
    
    \label{tab1}
\end{table}

As shown in Table \ref{tab1}, filtering the semantic segmentation score tensor by ACF significantly improves recognition performance. Compared to inputting the original score tensor, the use of ACF in SSRM increases the recognition accuracy by 3.43\% to 5.75\% on MIT-67 and 2.365\% to 3.705\% on SUN397, while reducing the number of Flops by 17.89G. Additionally, ChAM further improves the accuracy of SSRM with only a slight increase in Flops.

Next, we explore the effect of using ACF with different filter domain sizes on the performance of SSRM. Our results show that using a $2 \times 2$ ACF improves the accuracy of SSRM on MIT-67 and SUN397 by 1.12\% and 0.929\%, respectively, compared to using a $4 \times 4$ ACF. This suggests that while ACF mitigates the negative impact of semantic ambiguity, it may also cause the input tensor to lose some object information. Therefore, when selecting the filter domain size of ACF, both the loss of object information and the presence of semantic ambiguity should be considered. Notably, despite losing information, a $4 \times 4$ ACF still significantly enhances SSRM's accuracy by 3.43\% and 2.776\% on MIT-67 and SUN397 datasets, respectively. These results reaffirm that utilizing ACF to mitigate semantic ambiguity is significant for SSRM.

Furthermore, we pretrain the SSRM on the Places dataset \citep{r18}, achieving higher accuracy as shown in the last row of Table \ref{tab1}.

\subsubsection{Evaluation of different feature aggregation strategies}

Combining the label map generated by the semantic segmentation and the scene-related features generated by the SSRM and IFEM is essential for the Semantic Node Feature Aggregation Module. This module's output (i.e., two semantic feature sequences) directly affects the subsequent module's exploration for feature dependencies; therefore, preliminary experiments are necessary to identify the most appropriate aggregation strategy. To generate appropriate semantic feature sequences, we try a series of strategies to integrate the label map and scene-related features, as shown in Table \ref{tab2}.

In Table \ref{tab2}, $filtering$ represents using ACF with the specified kernel size to filter the semantic segmentation score tensor, in the same way as the disambiguation filtering process in SSRM. $Nearest$ means that the label map is interpolated to the specified size using nearest-neighbor interpolation, and $Bilinear$ means that the scene-related features are interpolated to the specified size using bilinear interpolation. For clarity in our presentation, we have numerically labeled each aggregation strategy. In this part, we first combine the output of SSRM and IFEM directly via Maximum and feed it to the classifier, using this result as the baseline performance, numbered as experiment 0.

\begin{table}[htbp]
    \small
    \centering
    \caption{Ablation results for different feature aggregation strategies (\%).}
    \begin{adjustbox}{width=0.8\textwidth}
    \begin{tabular}{ccccc}
    \hline
        Experiment & Score tensor & Label map & Scene features & MIT-67  \\ \hline
        0  & - & - & - & 88.731  \\ 
        1  & - & $Nearest$ to $14 \times 14$ & - & 89.851  \\ 
        2  & $2 \times 2$ $filtering$ & $Nearest$ to $14 \times 14$ & - & \textbf{90.746}  \\ 
        3  & $4 \times 4$ $filtering$ & $Nearest$ to $14 \times 14$ & - & 90.149  \\ 
        4  & - & - & $Bilinear$ to $224 \times 224$ & 89.776  \\ 
        5  & $2 \times 2$ $filtering$ & - & $Bilinear$ to $112 \times 112$ & 90.224  \\ 
        6  & $4 \times 4$ $filtering$ & - & $Bilinear$ to $56 \times 56$ & 90.299  \\ 
        7  & - & $Nearest$ to $56 \times 56$ & $Bilinear$ to $56 \times 56$ & 90.075  \\ 
        8  & $2 \times 2$ $filtering$ & $Nearest$ to $56 \times 56$ & $Bilinear$ to $56 \times 56$ & 90.672  \\ \hline
    \end{tabular}
    \end{adjustbox}
    \label{tab2}
\end{table}

As shown in Table \ref{tab2}, our method consistently outperforms the baseline regardless of the strategy used to aggregate the label map and scene-related features. This suggests that assigning scene features to objects and modeling long-range object co-occurrence via the Global-Local Dependency Module, results in more discriminative scene representations and enhanced recognition performance. Furthermore, upon experiments 1, 2, and 3; experiments 4, 5, and 6; and experiments 7 and 8, it can be concluded that applying the ACF to the score tensor consistently improves outcomes compared to the original score tensor. This finding aligns with the phenomenon presented in Table \ref{tab1}, further highlighting the effectiveness of ACF in semantic segmentation-based methods. Additionally, comparing experiment 2 with 3, it can be seen that a $4 \times 4$ filter domain produces slightly inferior results than a $2 \times 2$ filter domain. This observation again indicates the need to consider both ambiguity filtering and preservation of object information within scenes when processing the score tensor using ACF.

Upon comparing experiments 2, 5, and 8, it is evident that finer semantic label assignment to features (i.e., interpolating scene features to larger sizes) hurts final scene recognition. This phenomenon may be because the interpolation algorithm shifts the position of pixel-level features, increasing the risk of incorrect semantic label assignment. Further evidence supporting this observation comes from experiments 5 and 6, where the large interpolation range results in a $2 \times 2$ ACF filtered score tensor producing lower recognition than a $4 \times 4$ ACF filtered score tensor, a result that should have been the opposite. Fortunately, using un-interpolated scene features can expedite this module's processing. Therefore, we finally choose the configuration from experiment 2 to generate feature sequences. In this setup, the $2 \times 2$ ACF is first used to process the score tensor to generate an appropriate label map. This map is then interpolated to the size of scene features by nearest-neighbor interpolation. Finally, the label map and scene features are combined to generate the final semantic feature sequences.

\subsubsection{Evaluation of different Encoder combination methods}

In the Global-Local dependency Module, to ensure that the input features of the Decoder adequately preserve the complementary information of output features from two Encoders, we evaluate four different Encoder combination methods: Product, Concatenation, Addition, and element-wise Maximum. We conduct comparative experiments on the MIT-67 and SUN397 datasets, with results presented in Table \ref{tab3}. 

\begin{table}[htbp]
    \footnotesize
    \centering
    \caption{Ablation results for different Encoder combination methods (\%).}
    \begin{tabular}{lcc}
    \hline
        Method & MIT-67 & SUN397  \\ \hline
        Product & 89.925 & 75.965  \\ 
        Concatenation & 90.299 & 75.859  \\ 
        Addition & 90.373 & 75.965  \\ 
        Maximum & \textbf{90.746} & \textbf{76.153}  \\ \hline
    \end{tabular}
    \label{tab3}
\end{table}

The results in Table \ref{tab3} reveal that the element-wise Maximum method surpasses all other combination methods. To be specific, compared to the Concatenation method, the Maximum method requires fewer computational resources and achieves enhanced performance. In addition, compared to the Product and Addition methods, the Maximum method consumes the same computational resources while preserving the key information more cleanly, which allows the Decoder to decode dependencies in greater depth, ultimately resulting in superior scene recognition performance. Consequently, we finally choose the element-level Maximum to combine encoded features.

\subsubsection{Network components analysis}

In this part, we conduct a comprehensive network component analysis for the proposed method on the MIT-67 dataset. Initially, we construct a baseline model, which takes the output features from the Image Feature Extraction Module as input, and then applies average pooling and a fully connected layer for scene prediction (PlacesCNN). Subsequently, we evaluate the effect of four key components: PlacesCNN (baseline), Semantic Spatial Relation Module (SSRM), Encoder, and Decoder (Global-Local Dependency Module). Experimental results are presented in Table \ref{tab4}.

\begin{table}[htbp]
    \footnotesize
    \centering
    \caption{Ablation study of all components (\%).}
    \begin{tabular}{cccccc}
    \hline
        PlacesCNN & SSRM & Encoder & Decoder  & Accuracy  \\ \hline
        \checkmark & - & - & -  & 84.970\\ 
        \checkmark & \checkmark & - & -  & 88.731  \\ 
        \checkmark & \checkmark & \checkmark & -  & 90.075 \\ 
        \checkmark & \checkmark & \checkmark & \checkmark  & \textbf{90.746}  \\ \hline
        \multicolumn{4}{c}{Improvement Over Baseline (PlacesCNN)} & 5.776 \\ \hline
    \end{tabular}
    \label{tab4}
\end{table}

In Table \ref{tab4}, the best recognition accuracy is highlighted in bold, indicating an improvement of up to 5.776\% over the baseline, which strongly demonstrates the effectiveness of the proposed SpaCoNet in scene recognition. Furthermore, it can be concluded that each component of SpaCoNet contributes to the overall performance enhancement. More specifically, compared to the baseline, the accuracy is improved by 3.761\% after incorporating spatial contextual features extracted by SSRM into the network. This substantial improvement confirms the efficacy of employing a dedicated neural network to adaptively explore spatial information for assisting scene recognition. Furthermore, by assigning scene-related features to objects and applying the Global-Local Dependency module to capture long-range correlations among object features, we observe a further increase in accuracy by 2.015\%. This result suggests that distinguishing objects across different scenes using scene-specific features, coupled with mining long-range dependencies via attention mechanisms, significantly enhances the feature discriminability, facilitating more accurate scene representation.

In addition, we conduct ablation experiments on the Encoder-Decoder architecture within Global-Local Dependency Module, i.e., we try merging the outputs from two Encoders and directly feeding them into the classifier for scene recognition (bypassing the Decoder). The above process yields slightly inferior results compared to the full Encoder-Decoder configuration. This performance gap likely stems from the Decoder's critical role in bridging the modal discrepancies between the outputs of PlacesCNN and SSRM. The Decoder enables the network to more effectively exploit the complementary information between two Encoders, thereby resulting in superior scene recognition performance. This result demonstrates the rationality and superiority of the Encoder-Decoder architecture in  Global-Local Dependency Module.

\subsection{Qualitative visualization analysis}
In this subsection, we carry out a series of visualization experiments to qualitatively demonstrate and analyze the effectiveness of the proposed method. Initially, we conduct Heat map visualization based on Grad-CAM \citep{selvaraju2017GradCAM}, exploring how the proposed module adjusts the network's region of interest from an instance-level perspective. Subsequently, we employ t-SNE \citep{van2008tSNE} to visualize all dataset samples, assessing the impact of our method on the network's discriminative ability from a dataset-level perspective.

\subsubsection{Instance-level visualization}
\begin{figure}[htbp]
    \centering
    \includegraphics[width=0.9\textwidth]{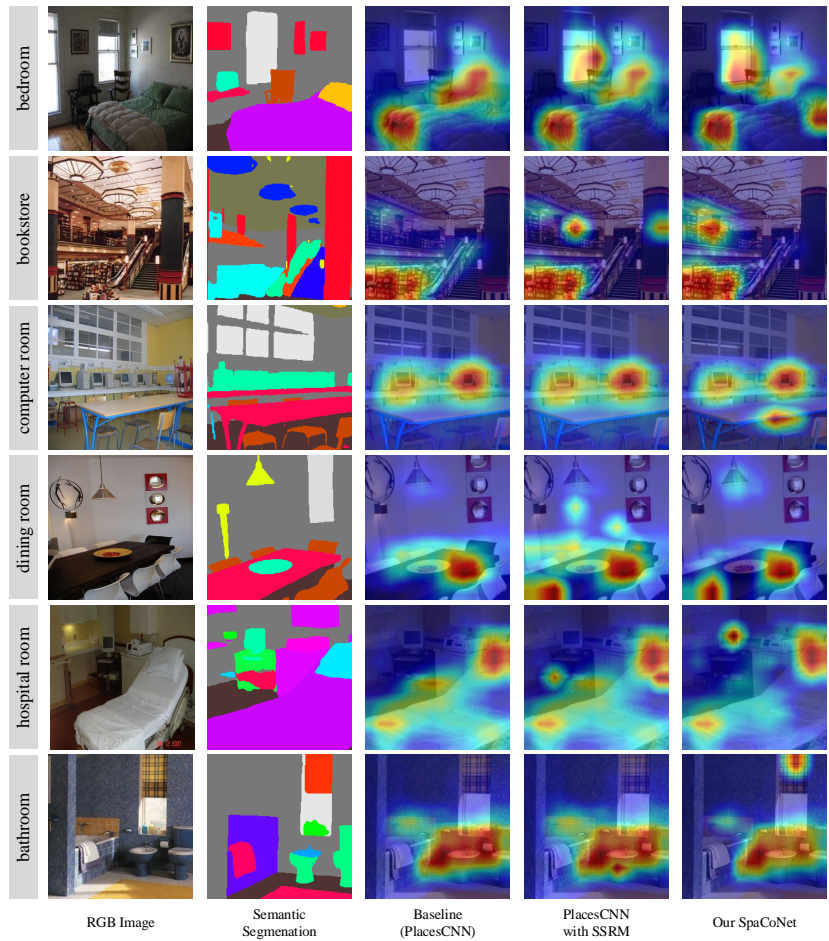}
    \caption{Heat maps generated using Grad-CAM for various images. The first two columns showcase RGB images from the MIT-67 test sets, accompanied by their semantic segmentation results (colored). The next three columns showcase heat maps generated from three methods: Baseline (PlacesCNN), PlacesCNN with SSRM (spatial relation), and our SpaCoNet (spatial relation with object co-occurrence). Regions more emphasized by the network will appear warmer in heat maps. Note that when performing visualization for the latter two methods, which each have two branches, we follow the feature aggregation strategy used for two Encoders (see Eq. \ref{eq_maximum} for details), using the Maximum strategy to merge the heat maps from two branches into the final heat map.}
    \label{FIG_heat_map}
\end{figure}

\begin{figure}[htbp]
    \centering
    \includegraphics[width=0.9\textwidth]{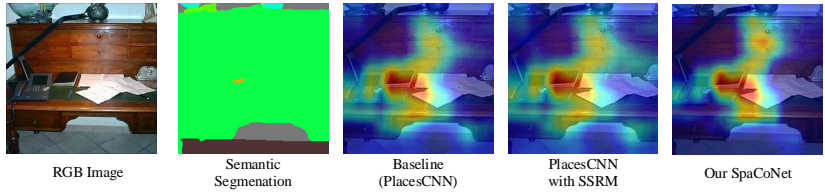}
    \caption{Limitations of the proposed method. In some cases, when semantic segmentation fails to effectively segment objects within a scene, our method's capacity for adjusting the network's region of interest is severely limited.}
    \label{FIG_limitation}
\end{figure}

In this part, we use Grad-CAM to visualize the proposed module's effect on the network's discriminative region. Fig. \ref{FIG_heat_map} and Fig. \ref{FIG_limitation} display the visualization results in the following sequence, from left to right: the original color image, the semantic segmentation label map (colored), the heat map based on the baseline (PlacesCNN), the heat map based on the combination of PlacesCNN and SSRM (exploring spatial relation), and the heat map based on the proposed SpaCoNet (further exploring object co-occurrence).

\paragraph{\textbf{Advantages}}
As demonstrated in Fig. \ref{FIG_heat_map}, the proposed modules actively adjust the network's focus regions, enabling the network to more effectively utilize objects within images for scene prediction. For example, in the bedroom scene, the baseline primarily focuses on "bed" (third column); after introducing spatial relation exploration (fourth column), the network further focuses on "window"; in the SpaCoNet (fifth column), attention also extends to "pillow." Meanwhile, in some cases, merely exploring spatial relations is sufficient to represent the scene. For instance, in the bookstore scene, exploring spatial relations drives the network to perceive the more distant "bookshelves" upstairs (fourth column). This suggests that spatial relationships furnish additional complementary information for the network to concentrate on finer scene details. However, since no more significantly distinctive areas exist, further exploring object co-occurrence (fifth column) just broadens the network's focus on "bookshelves."

Furthermore, in certain cases, exploring spatial relations provides limited benefit to the network; under such conditions, integrating scene features to explore object co-occurrence is more helpful. For example, in the computer room scene, merely exploring spatial relations (fourth column) fails to effectively adjust the focal areas of the baseline (third column). This ineffectiveness may be because "table" are common in multiple scenes (such as dining rooms). In contrast, SpaCoNet (fifth column) assigns scene-related features to objects and models object co-occurrence, enabling the network to recognize unique "table" features in the computer room, thus further identifying  "table" as a key area for scene prediction. Similarly, in the hospital room scene, since "computer" is common in others (like the computer room), the third and fourth columns do not consider "computer" as a contributing factor for scene prediction. Conversely, SpaCoNet (fifth column), by integrating scene features to explore object co-occurrence, further considers "computer" as an important element for scene prediction. These examples fully demonstrate the superiority of our method that combines scene features to model object co-occurrence for scene recognition.

\paragraph{\textbf{Limitations}}
\label{Limitations}
The experiments discussed above confirm the efficacy of SpaCoNet for scene recognition. However, it is important to acknowledge that SpaCoNet's performance significantly depends on the quality of the semantic segmentation applied to the input scene.

For instance, in Fig. \ref{FIG_limitation}, the semantic segmentation technique used fails to identify most objects within this scene. The regions of interest highlighted in columns 4 and 5 are almost the same as the baseline (column 3), suggesting that our proposed module contributes minimally to scene recognition under these circumstances. This limitation arises because, without adequate semantic prior knowledge, SpaCoNet struggles to effectively model spatial relation and object co-occurrence. Fortunately, despite the absence of prior knowledge, our method does not negatively affect the recognition performance. This may be attributed to our two-stage training strategy, in which PlacesCNN (baseline) weights are frozen during the second training stage, thereby maintaining its fundamental recognition ability.

\subsubsection{Dataset-level visualization}
Considering the limited number of samples that can be displayed in instance-level visualization experiments, we further conduct dataset-level visualization analysis to comprehensively demonstrate the generalization of our method from a broader perspective. 

\begin{table}[htbp]
    \footnotesize
    \centering
    \caption{Ablation study of feature visualization (\%).}
    \begin{tabular}{lcccc}
    \hline
        Method & MIT-67 & Places\_7 & Places\_14 & SUN397  \\ \hline
        Baseline (PlacesCNN) & 84.970 & 93.0 & 87.643 & 73.129  \\ 
        Our SpaCoNet & 90.746 & 94.286 & 89.714 & 76.153  \\ \hline
    \end{tabular}
    \label{tab5}
\end{table}

\begin{figure}[htbp]
    \centering
    \includegraphics[width=\textwidth]{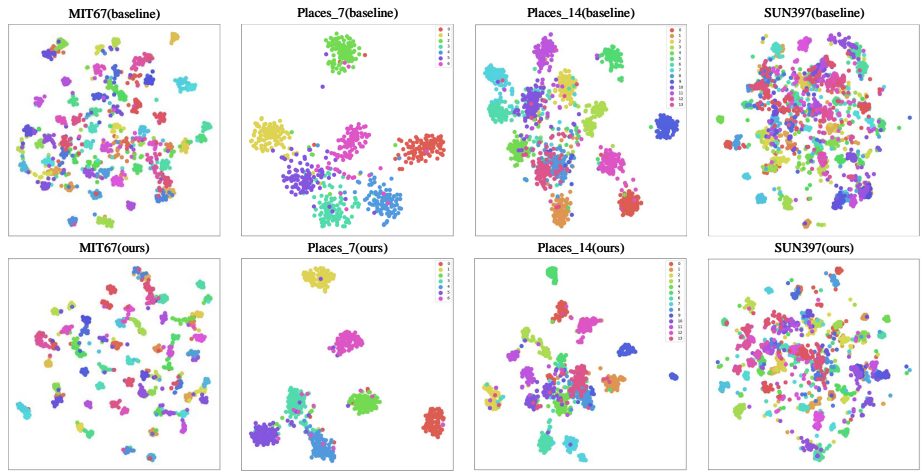}
    \caption{Feature distributions of scene categories (different colors representing different categories).}
    \label{Fig7}
\end{figure}

Specifically, we evaluate the proposed SpaCoNet and the baseline (PlacesCNN) on the test sets of MIT-67, Places\_7, Places\_14, and SUN397. The results of these evaluations are presented in Table \ref{tab5}. Furthermore, we extract the features that will be fed into the classifier and use t-SNE to visualize them by plotting their 2-dimensional representations in Fig. \ref{Fig7}, in which each point represents an image and points with the same color indicate images from the same category. The first row of Fig. \ref{Fig7} displays the feature visualization from PlacesCNN, while the second row displays the feature visualization from our SpaCoNet.

As shown in Fig. \ref{Fig7}, compared to the baseline feature distribution, the proposed method results in a more clustered distribution of scene features from the same category and a more dispersed distribution of scene features from different categories. This phenomenon is particularly pronounced on less-categorical Places\_7 and Places\_14 datasets, in which only fewer different points are mixed. Furthermore, even in the MIT-67 and SUN397 datasets with numerous categories, it is still evident that our method enhances the boundaries between different scene categories and reduces the differences between the same categories. These results demonstrate that simultaneously modeling spatial relationships and long-range dependencies among objects significantly improves the discriminative ability of scene features, resulting in superior scene recognition performance (refer to Table \ref{tab5}).

\subsection{Comparison with state-of-the-art methods}

To demonstrate the competitive performance of the proposed method, we compare SpaCoNet with state-of-the-art methods on multiple datasets: Places\_14, Places\_7, reduced SUN RGB-D, MIT-67, and SUN397. When validating on reduced SUN RGB-D, the model is pretrained on Places\_7 without retraining to compare the generalization. The performance comparison results are presented in Tables \ref{tab6} and \ref{tab7}. Meanwhile, we compare the computational complexity of our SpaCoNet with existing methods using Flops as the metric, with results shown in Table \ref{tab8}.

\begin{table}[htbp]
    \footnotesize
    \centering
    \caption{State-of-the-art results on Places\_14, Places\_7, and SUN RGB-D dataset (\%).}
    \begin{adjustbox}{width=0.75\textwidth}
    \begin{tabular}{ccccc}
    \hline
        Approaches  & Network Input Size & Places\_14 & Places\_7 & SUN RGB-D \\ \hline
        Word2Vec \citep{r36}  & 224 $\times$ 224 & 83.7 & - & - \\ 
        Deduce \citep{r23} & 224 $\times$ 224 & - & 88.1 & 70.1  \\ 
        BORM-Net \citep{r25} & 224 $\times$ 224 & 85.8 & 90.1  & 72.1  \\ 
        OTS-Net \citep{r20} & 224 $\times$ 224 & 85.9 & 90.1  & 70.6  \\
        AGCN \citep{zhou2023attentional}   & 224 $\times$ 224  & 86.0 & 91.7  & -  \\ 
        CSRRM \citep{song2023srrm} & 224 $\times$ 224 & 88.714 & 93.429  & 75.349  \\ 
        Our SpaCoNet & 224 $\times$ 224 & \textbf{89.714} & \textbf{94.286}  & \textbf{77.853}  \\ \hline
    \end{tabular}
    \end{adjustbox}
    \label{tab6}
\end{table}

As demonstrated in Table \ref{tab6} and \ref{tab7}, the proposed SpaCoNet outperforms most existing methods in performance. For example, methods \citep{r17,r19,r22,zhou2023attentional,EMLELM_ESWA,wang2023adaptive} rely on pre-trained neural networks to identify discriminative regions, combining strategies such as feature encoding or graph convolution for scene recognition. These methods are constrained by the pretrained network's feature discriminative ability. In contrast, SpaCoNet enhances scene recognition by integrating object information, leveraging object prior knowledge to achieve superior results. In addition, SpaCoNet outperforms some multi-branch-based methods \citep{r16,r29,kbs2021content,MVML_ESWA,DeepScene_ESWA,r38,NEM_MVIP024}, which aim to obtain multi-scale scene information for scene recognition. This phenomenon again highlights the effectiveness of using object information as an additional source.

\begin{table}[!ht]
    \centering
    \caption{State-of-the-art results on MIT-67 and SUN397 datasets (\%).}
    \begin{adjustbox}{width=0.65\textwidth}
    \begin{tabular}{cccc}
    \hline
        Approaches  & Network Input Size  & MIT-67  & SUN397  \\ \hline
        Dual CNN-DL \citep{r17} & 224 $\times$ 224 & 76.56 & 70.13  \\ 
        NNSD + ICLC \citep{r19} & 224 $\times$ 224 & 84.3 & 64.78 \\
        MVML-LSTM \citep{MVML_ESWA} & 256 $\times$ 256  & 80.52  & 63.02  \\  
        Multi-Resolution CNNs \citep{r16} & 336 $\times$ 336 & 86.7 & 72.0 \\ 
        LGN \citep{r22}   & 448 $\times$ 448  & 88.06  & 74.06  \\ 
        Scene Essence \citep{r6}   & 224 $\times$ 224  & 83.92  & 68.31  \\ 
        MRNet \citep{r29}   & 448 $\times$ 448  & 88.08  & 73.98  \\
        DeepScene \citep{DeepScene_ESWA} & 224 $\times$ 224  & 71.0  & -  \\ 
        EML-ELM \citep{EMLELM_ESWA} & 224 $\times$ 224  & 87.1  & -  \\ 
        SDO \citep{r8}   & 224 $\times$ 224  & 86.76  & 73.41  \\ 
        SAS-Net \citep{r21}   & 224 $\times$ 224  & 87.1  & 74.04  \\ 
        CCF \citep{kbs2021content}   & 512 $\times$ 512  & 87.3  & -  \\
        AdaNFF \citep{r45}  & 256 $\times$ 256  & - & 74.18  \\ 
        ARG-Net \citep{r24}   & 448 $\times$ 448  & 88.13  & 75.02  \\
        FCT \citep{r38} & 224 $\times$ 224  & 89.17  & 76.06  \\ 
        NEM \citep{NEM_MVIP024}   & 224 $\times$ 224  & 89.17  & 76.06  \\
        ALR-Net \citep{wang2023adaptive}   & 448 $\times$ 448  & 88.37  & 74.24  \\ 
        CSRRM \citep{song2023srrm}   & 224 $\times$ 224  & 88.731  & 75.476*  \\ 
        Our SpaCoNet  & 224 $\times$ 224  & \textbf{90.746}  & \textbf{76.153}  \\ \hline
        \multicolumn{4}{l}{\ {*} denotes the CSRRM results as reproduced in our experiments.}
    \end{tabular}
    \end{adjustbox}
    \label{tab7}
\end{table}

Furthermore, compared to methods \citep{r8,r23,r21,r24,r20,r25,r36} that also incorporate object information for scene recognition, SpaCoNet demonstrates enhanced performance, highlighting the effectiveness of simultaneously exploiting spatial relation and co-occurrence among objects. To be specific, methods \citep{r22,r24} perform scene recognition by modeling spatial metric relationships among object regions; however, spatial relationships within scenes are not limited to metric relationships. In contrast, SpaCoNet first decouples spatial information from the scene and then employs a specialized branch to thoroughly model spatial features. This process enables more effective adaptation to complex and variable spatial layouts, resulting in superior scene recognition. In addition, while some statistical methods \citep{r23,r25,r8} also assist scene recognition by modeling object co-occurrence, they overlook the challenges posed by discriminative objects that coexist across different scenes. In contrast, SpaCoNet assigns scene-related features to individual objects, thus distinguishing objects with identical semantics across scenes. Then, SpaCoNet models object co-occurrence through the attention mechanism, yielding more advanced recognition performance.

Moreover, compared to methods \citep{zhou2023attentional,r29,r38,MVML_ESWA,NEM_MVIP024} that also utilize the attention mechanism for scene recognition, SpaCoNet employs semantic prior knowledge to initially assign scene-related features to objects, subsequently exploring the correlations among these object features through the attention mechanism. This process effectively exploits the attention mechanism's advantage in modeling long-range dependencies among sequential features, thereby achieving superior recognition performance.

\begin{table}[htbp]
    \footnotesize
    \centering
    \caption{Computational complexity comparison.}
    \begin{adjustbox}{width=1\textwidth}
    \begin{tabular}{ccccccc}
        \hline
            Approaches & Flops (G) & MIT-67 (\%) & SUN397 (\%) & Places\_7 (\%) & Places\_14 (\%) & SUN RGB-D (\%) \\ \hline
            SASLM \citep{wang2024single} & 26.97 & 88.58 & 73.96 & - & -  & -  \\ 
            DeepScene \citep{DeepScene_ESWA} & 19.91 & 71.0 & - & - & -  & -  \\ 
            OTS-Net \citep{r20} & 0.214 & - & - & 90.1 & 85.9  & 70.6  \\ 
            AGCN \citep{zhou2023attentional} & 54.746 & - & - & 91.7 & 86.0  & -  \\ 
            BORM-Net \citep{r25} & 138.011 & - & - & 90.1 & 85.8  & 71.1  \\ 
            SpaCoNet & 48.705 & 90.746 & 76.153 & 94.286 & 89.714  & 77.853  \\ \hline
        \end{tabular}
    \end{adjustbox}
    \label{tab8}
\end{table}

As shown in Table \ref{tab8}, SpaCoNet demonstrates a competitive balance between computational efficiency and recognition accuracy. With 48.705G Flops, it is more efficient than BORM-Net (138.011G) \citep{r25} and AGCN (54.746G) \citep{zhou2023attentional} while outperforming them across all shared datasets. Although SpaCoNet requires more computation than some methods \citep{wang2024single,DeepScene_ESWA,r20}, it achieves significantly higher accuracy on multiple datasets.

Despite not being the most computationally efficient model, SpaCoNet's consistently superior performance across all evaluated datasets underscores its effectiveness in scene recognition tasks. This balanced approach positions SpaCoNet as an ideal choice for applications where high accuracy is paramount, without imposing excessive computational demands.

Overall, all the above comparisons confirm the superiority and generalization of the proposed SpaCoNet for indoor scene recognition.

\section{Conclusions}

In this paper, we propose a framework, SpaCoNet, to simultaneously model the \textbf{Spa}tial relation and \textbf{Co}-occurrence of objects for indoor scene recognition. 

Initially, we construct the Semantic Spatial Relation Module (SSRM) to specifically model spatial contextual relationships within scenes. Visualization experiments demonstrate that using dedicated modules to implicitly model spatial information allows the network to focus on finer scene details. In SSRM, an Adaptive Confidence Filter (ACF) is designed to mitigate the impact of errors in semantic segmentation results. Ablation studies for SSRM confirm the ACF's significant contribution to improving scene recognition performance. ACF offers a new solution for mitigating semantic ambiguity in semantic-guided methods.

Additionally, to distinguish objects with identical semantics across different scenes while exploring object co-occurrence, we construct the Semantic Node Feature Aggregation Module. This module allocates scene-related features to objects and reformulates them as semantic node feature sequences, enabling SpaCoNet to use scene-related features to differentiate objects with identical semantics. Furthermore, we design the Global-Local Dependency module to explore the long-range correlation among semantic node features and global features. Qualitative analysis reveals that by integrating scene-related features to explore object co-occurrence, SpaCoNet effectively leverages abundant object information—including cross-scene co-existing objects—resulting in more discriminative scene representations.

Compared to existing scene recognition methods, SpaCoNet demonstrates highly competitive recognition performance on multiple public datasets, without imposing excessive computational demands. This outcome validates the effectiveness of simultaneously modeling spatial relations and object co-occurrence. Overall, our framework offers new research avenues in semantic-guided scene recognition, showcasing the potential of dedicated branches for spatial information modeling and scene-related feature integration to capture object co-occurrence.

However, as mentioned in Section \ref{Limitations}, the performance of the proposed method is constrained by the semantic segmentation technique employed. Specifically, the segmentation technique used in this study is capable of segmenting 150 semantic objects; however, it does not cover all objects present in scenes, which is one of the reasons why we incorporate global features into the semantic feature sequence. Two potential strategies can be considered to address this issue. The first strategy is to train segmentation techniques that can effectively segment more semantic objects. However, this strategy necessitates significant effort to annotate the dataset, making it a resource-intensive endeavor. Alternatively, the second strategy is to use an unsupervised or semi-supervised approach to enable the network to autonomously recognize objects within the scene, which is a direction that we intend to pursue in the future.

Finally, we believe the proposed method can inspire new research avenues beyond scene recognition. Specifically, our exploration of spatial relations reveals that for complex key information, decoupling it from the input and designing dedicated branches to extract high-level feature representations is a feasible resolution. This could be adapted to address similar challenges in other tasks. Additionally, the inter-class similarity challenges, arising from the semantic consistency of discriminative objects across different scenes, are common in various fields, such as remote sensing scene classification and multi-label scene recognition. We anticipate that our approach of integrating scene-related features to model object co-occurrence will provide fresh insights into these areas.

\section*{Acknowledgment}
This work was jointly supported by the Key Development Program for Basic Research of Shandong Province under Grant ZR2019ZD07, the National Natural Science Foundation of China-Regional Innovation Development Joint Fund Project under Grant U21A20486, the Fundamental Research Funds for the Central Universities under Grant 2022JC011.



\bibliographystyle{model5-names.bst}\biboptions{authoryear}
\bibliography{refs.bib}





\end{document}